\documentclass[conference]{IEEEtran}

\usepackage[utf8]{inputenc}
\usepackage[T1]{fontenc}
\usepackage{amsmath,amssymb,amsfonts}
\usepackage{graphicx}
\usepackage{booktabs}
\usepackage{multirow}
\usepackage{cite}
\usepackage{hyperref}
\usepackage{xcolor}
\usepackage{float}
\usepackage{stfloats}

\hypersetup{
    colorlinks=true,
    linkcolor=black,
    citecolor=black,
    urlcolor=blue
}

\begin{document}

\title{Visual Chart Representations for Cryptocurrency\\Regime Prediction: A Systematic Deep Learning Study}

\author{\IEEEauthorblockN{Dustin M. Haggett}
\IEEEauthorblockA{Department of Electrical and Computer Engineering\\
Stevens Institute of Technology\\
Hoboken, NJ, USA\\
dhaggett@stevens.edu\\
November 29, 2025}}

\maketitle

\begin{abstract}
Technical traders have long relied on visual analysis of candlestick charts to identify market patterns and predict price movements. While deep learning has achieved remarkable success in image classification, its application to financial chart images remains underexplored. This paper presents a systematic study comparing different visual representations for cryptocurrency regime prediction. We evaluate three image encoding methods (raw candlestick charts, Gramian Angular Fields, and multi-channel GAF), five chart component configurations, four neural network architectures (CNN, ResNet18, EfficientNet-B0, and Vision Transformer), and the impact of ImageNet transfer learning. Through eight controlled experiments on Bitcoin, Ethereum, and S\&P 500 data spanning 2018--2024, we identify optimal configurations for visual regime classification. Our results show that a simple 4-layer CNN on raw candlestick charts achieves 0.892 AUC-ROC, outperforming larger pretrained models. Surprisingly, simpler representations (price-only charts, 128$\times$128 resolution) consistently outperform more complex alternatives. We provide interpretability analysis using GradCAM and demonstrate that transfer learning improves performance by 4--16\% despite the domain gap between natural images and financial charts.
\end{abstract}


\section{Introduction}

Technical analysis has been a cornerstone of financial trading for centuries, with Japanese candlestick charting dating back to 18th-century rice markets \cite{nison2001japanese}. Traders visually analyze chart patterns---such as head and shoulders, double tops, and engulfing patterns---to predict future price movements. This visual approach to market analysis suggests that predictive information may be encoded in the spatial arrangement of price data.

The success of convolutional neural networks (CNNs) in image classification tasks \cite{krizhevsky2012imagenet} raises a natural question: \textit{Can deep learning models learn to ``see'' predictive patterns in financial charts the way human traders do?}

Recent work has explored this direction with promising results. Kusuma et al. \cite{kusuma2019using} achieved 92\% accuracy predicting stock movements from candlestick images using CNNs, while Chen and Tsai \cite{chen2020encoding} proposed GAF-CNN, encoding time series as Gramian Angular Field images for pattern classification. However, several fundamental questions remain unanswered regarding the optimal approach for visual financial prediction.

First, it remains unclear how different visual representations compare---specifically whether raw chart images or mathematical encodings like Gramian Angular Fields better preserve predictive information. Second, the impact of chart visual components such as volume bars, moving averages, and Bollinger Bands on model performance has not been systematically studied. Third, while Vision Transformers have revolutionized computer vision, their application to financial charts remains unexplored. Fourth, the effectiveness of transfer learning from ImageNet to this specialized visual domain is unknown. Finally, whether models trained on one asset generalize to others has not been investigated.

This paper addresses these questions through a systematic experimental study comprising eight controlled experiments. We make several contributions to the literature: a comprehensive comparison of three image encoding methods for financial time series, ablation studies on chart visual components and their predictive value, the first comparison of CNN versus Vision Transformer architectures on candlestick chart images, evaluation of transfer learning effectiveness for financial chart classification, GradCAM interpretability analysis revealing what visual features CNNs learn from charts, and an open-source implementation with trained models for reproducibility.

\section{Related Work}

\subsection{Deep Learning for Financial Prediction}

Deep learning has been extensively applied to financial time series prediction over the past decade. Recurrent architectures, particularly Long Short-Term Memory networks \cite{hochreiter1997long}, have been popular for capturing temporal dependencies in price sequences. Fischer and Krauss \cite{fischer2018deep} demonstrated that LSTMs could outperform traditional methods on stock market prediction tasks. More recently, Transformer-based models have shown promise for financial forecasting \cite{wu2020deep}, leveraging self-attention mechanisms to capture long-range dependencies. However, these approaches process numerical sequences rather than visual representations, potentially missing the spatial patterns that human traders perceive in charts.

\subsection{Chart-Based Deep Learning}

The idea of treating financial data as images emerged from the observation that traders make decisions based on visual chart patterns rather than raw numbers. Kusuma et al. \cite{kusuma2019using} pioneered this approach by converting candlestick data to images and training CNN, ResNet, and VGG networks, achieving up to 92\% accuracy on Taiwanese stocks. Their work demonstrated that CNNs could learn meaningful features from chart visualizations. Subsequently, Hung and Chen \cite{hung2021dpp} proposed the Deep Predictor for Price model using CNN autoencoders on candlestick sub-charts, showing that unsupervised feature learning could capture relevant chart patterns.

\subsection{Gramian Angular Fields}

Wang and Oates \cite{wang2015encoding} introduced Gramian Angular Fields as a method to encode time series as images while preserving temporal dependencies. The approach represents time series values as angular coordinates in a polar coordinate system and computes pairwise trigonometric sums or differences to form a matrix visualization. This encoding theoretically preserves temporal correlations that might be lost in standard image representations. Chen and Tsai \cite{chen2020encoding} applied GAF-CNN to candlestick pattern recognition, achieving 90.7\% accuracy on eight classical patterns, suggesting that mathematical encodings could be effective for financial applications.

\subsection{Vision Transformers for Time Series}

Vision Transformers \cite{dosovitskiy2020image} have revolutionized computer vision by applying self-attention mechanisms to image patches, achieving state-of-the-art results on numerous benchmarks. Recent work has begun exploring ViT for time series applications. Zeng et al. \cite{zeng2023pixels} used spectrograms with ViT for forecasting, demonstrating that the architecture could capture relevant patterns in visual representations of temporal data. However, direct application of Vision Transformers to candlestick chart images remains unexplored, representing a gap this paper addresses.

\subsection{Research Gap}

While prior work has demonstrated the feasibility of chart-based prediction, no systematic study has compared different image encoding methods on the same prediction task, evaluated the impact of chart visual components, compared CNN versus Transformer architectures for this domain, or assessed transfer learning effectiveness. This paper fills these gaps through controlled experimentation.

\section{Methodology}

\subsection{Problem Formulation}

We formulate regime prediction as a binary image classification task. Given a chart image $\mathbf{X} \in \mathbb{R}^{H \times W \times C}$ representing the past $N$ days of price data, the goal is to predict the regime label $y \in \{0, 1\}$ defined as:

\begin{equation}
y = \begin{cases}
1 & \text{if } \frac{P_{t+k} - P_t}{P_t} > \tau \quad \text{(Bull regime)} \\
0 & \text{otherwise} \quad \text{(Bear regime)}
\end{cases}
\end{equation}

where $P_t$ is the closing price at time $t$, $k$ is the forward horizon (7 days in our experiments), and $\tau$ is the return threshold (2\%). This formulation transforms the continuous prediction problem into a classification task that can leverage advances in image recognition.

\subsection{Image Encoding Methods}

We compare three methods for converting OHLCV (Open, High, Low, Close, Volume) data to images, each with different theoretical properties.

\subsubsection{Raw Candlestick Charts}

The first encoding uses standard candlestick visualizations rendered using the \texttt{mplfinance} library. Each candle encodes four price points---open, high, low, and close---through its body and wicks. Green coloring indicates bullish days where the close exceeded the open, while red indicates bearish days. This representation directly mirrors what human traders analyze, potentially allowing CNNs to learn similar visual patterns. Optional overlays include volume bars displayed as a subplot, simple moving averages rendered as lines, and Bollinger Bands showing volatility envelopes.

\subsubsection{Gramian Angular Field (GAF)}

The second encoding follows Wang and Oates \cite{wang2015encoding} to transform the closing price series $X = \{x_1, ..., x_n\}$ into an image through a polar coordinate transformation. The process begins by rescaling values to the range $[-1, 1]$ using min-max normalization. These rescaled values are then converted to angular coordinates through the arccosine function: $\phi_i = \arccos(\tilde{x}_i)$. Finally, the Gramian Angular Summation Field is computed as:

\begin{equation}
G_{ij}^{GASF} = \cos(\phi_i + \phi_j)
\end{equation}

This encoding theoretically preserves temporal dependencies through the matrix structure, where each cell represents the relationship between two time points.

\subsubsection{Multi-Channel GAF}

The third encoding extends GAF to multiple channels by computing separate GASF matrices for different price components and stacking them as RGB channels:

\begin{equation}
\mathbf{G}_{RGB} = \text{stack}(G^{open}, G^{range}, G^{close})
\end{equation}

This approach aims to preserve more information from the OHLCV data by encoding open prices, the high-low range, and close prices in separate color channels.

\subsection{Chart Component Configurations}

To investigate whether additional visual elements improve predictions, we evaluate five configurations of increasing complexity. The simplest configuration shows only price candlesticks without any additional indicators. Subsequent configurations progressively add volume bars as a subplot, 7-day and 25-day simple moving averages overlaid on the price chart, and finally Bollinger Bands computed with a 20-day window and 2 standard deviations. The most complex configuration includes all elements simultaneously.

\subsection{Model Architectures}

We evaluate four neural network architectures spanning different design philosophies and parameter counts.

\subsubsection{Simple CNN}

Our baseline architecture is a four-block convolutional network with approximately 422,000 parameters. Each block consists of a convolutional layer followed by batch normalization, ReLU activation, and max pooling, with channel dimensions progressing from 3 to 32, 64, 128, and finally 256. The classifier head uses adaptive average pooling followed by two fully connected layers with dropout regularization. This architecture is designed to be simple enough to train from scratch on limited data while having sufficient capacity to learn relevant features.

\subsubsection{ResNet18}

We include the standard ResNet18 architecture \cite{he2016deep} with 11.2 million parameters, modified only in the final layer for binary classification. ResNet's skip connections enable training of deeper networks and have proven effective across numerous vision tasks. We evaluate both random initialization and ImageNet pretraining.

\subsubsection{EfficientNet-B0}

EfficientNet-B0 \cite{tan2019efficientnet} provides approximately 4.0 million parameters with a carefully balanced architecture designed for efficiency. Its compound scaling approach optimizes the trade-off between depth, width, and resolution, making it an interesting comparison point between our simple CNN and the larger ResNet.

\subsubsection{Vision Transformer (ViT-Tiny)}

To compare convolutional and attention-based approaches, we include ViT-Tiny \cite{dosovitskiy2020image} with 5.5 million parameters and a 16$\times$16 patch size. Vision Transformers process images as sequences of patches, potentially capturing different patterns than CNNs.

\subsection{Training Configuration}

All models are trained using the AdamW optimizer with a weight decay of $10^{-4}$ and an initial learning rate of $10^{-3}$. We use binary cross-entropy loss with logits, applying class weighting to handle the imbalanced distribution of bull and bear regimes. The learning rate is reduced by a factor of 0.5 when validation loss plateaus for 3 consecutive epochs, and training terminates early if no improvement is observed for 10 epochs. All experiments use a batch size of 32. Importantly, we determine the optimal classification threshold on the validation set by maximizing the F1 score, rather than using the default 0.5 threshold, which proved critical for handling class imbalance.

\section{Experimental Setup}

\subsection{Datasets}

We use daily OHLCV data downloaded from Yahoo Finance for three assets spanning January 2018 through December 2024. Bitcoin (BTC-USD) serves as our primary cryptocurrency, with Ethereum (ETH-USD) providing a second cryptocurrency for generalization testing. The S\&P 500 ETF (SPY) represents traditional equities for cross-asset comparison. Each dataset comprises 500 samples after applying the lookback and forward horizon requirements. The Bitcoin dataset exhibits approximately 40.8\% bull regimes under our 2\% threshold definition, indicating moderate class imbalance that necessitates careful handling during training and evaluation.

\begin{table}[h]
\centering
\caption{Dataset Statistics}
\label{tab:datasets}
\begin{tabular}{lccc}
\toprule
\textbf{Asset} & \textbf{Period} & \textbf{Samples} & \textbf{Bull \%} \\
\midrule
Bitcoin (BTC-USD) & 2018--2024 & 500 & 40.8\% \\
Ethereum (ETH-USD) & 2018--2024 & 500 & varies \\
S\&P 500 (SPY) & 2018--2024 & 500 & varies \\
\bottomrule
\end{tabular}
\end{table}

\subsection{Train/Validation/Test Split}

We employ time-series aware splitting to prevent look-ahead bias, which would occur if future data influenced predictions for earlier time points. The first 70\% of samples chronologically form the training set, the next 15\% comprise the validation set used for hyperparameter tuning and early stopping, and the final 15\% constitute the held-out test set for final evaluation. This approach ensures that all test predictions are made using only historically available information.

\subsection{Evaluation Metrics}

We report three metrics to comprehensively assess model performance. Accuracy measures overall classification correctness but can be misleading with imbalanced classes. The F1 score provides the harmonic mean of precision and recall, offering a balanced view of performance on both classes. Most importantly, AUC-ROC (Area Under the Receiver Operating Characteristic curve) measures discrimination ability independent of threshold selection. We emphasize AUC-ROC as our primary metric because it captures how well the model ranks bull regimes higher than bear regimes, regardless of the specific decision threshold chosen.

\subsection{Experiments}

We conduct eight experiments, each varying one factor while controlling others to isolate its effect. The baseline experiment establishes performance with a simple CNN on raw candlestick charts. Subsequent experiments systematically vary image encoding method, chart component configuration, historical lookback window (14, 30, 60, or 90 days), image resolution (64$\times$64, 128$\times$128, or 224$\times$224 pixels), model architecture, transfer learning strategy, and finally the asset being predicted.

\section{Results}

\subsection{Experiment 1: Baseline}

Our baseline configuration uses a simple CNN on raw candlestick charts with a 30-day lookback window and 224$\times$224 resolution. As shown in Table~\ref{tab:baseline}, the model achieves 49.3\% accuracy, 0.367 F1 score, and 0.760 AUC-ROC. While the accuracy appears poor, the AUC-ROC of 0.760 demonstrates that the model discriminates between bull and bear regimes significantly better than random chance (0.5), validating the viability of the visual approach. The discrepancy between accuracy and AUC-ROC highlights the importance of threshold selection with imbalanced classes.

\begin{table}[h]
\centering
\caption{Baseline Results}
\label{tab:baseline}
\begin{tabular}{lc}
\toprule
\textbf{Metric} & \textbf{Test Score} \\
\midrule
Accuracy & 49.3\% \\
F1 Score & 0.367 \\
AUC-ROC & 0.760 \\
Average Precision & 0.877 \\
\bottomrule
\end{tabular}
\end{table}

\begin{figure}[h]
\centering
\includegraphics[width=\columnwidth]{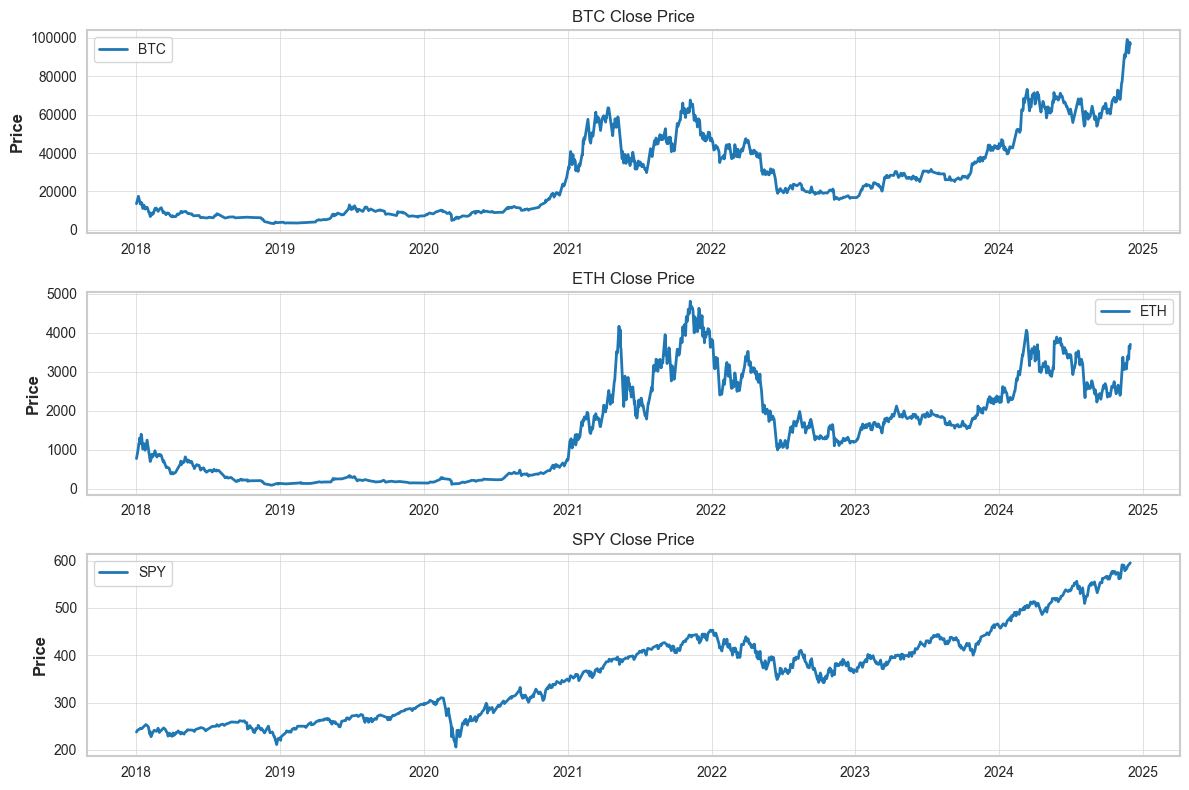}
\caption{Price history for Bitcoin, Ethereum, and S\&P 500 (2018--2024). Cryptocurrency markets exhibit substantially higher volatility than traditional equities.}
\label{fig:price_history}
\end{figure}

\begin{figure}[h]
\centering
\includegraphics[width=\columnwidth]{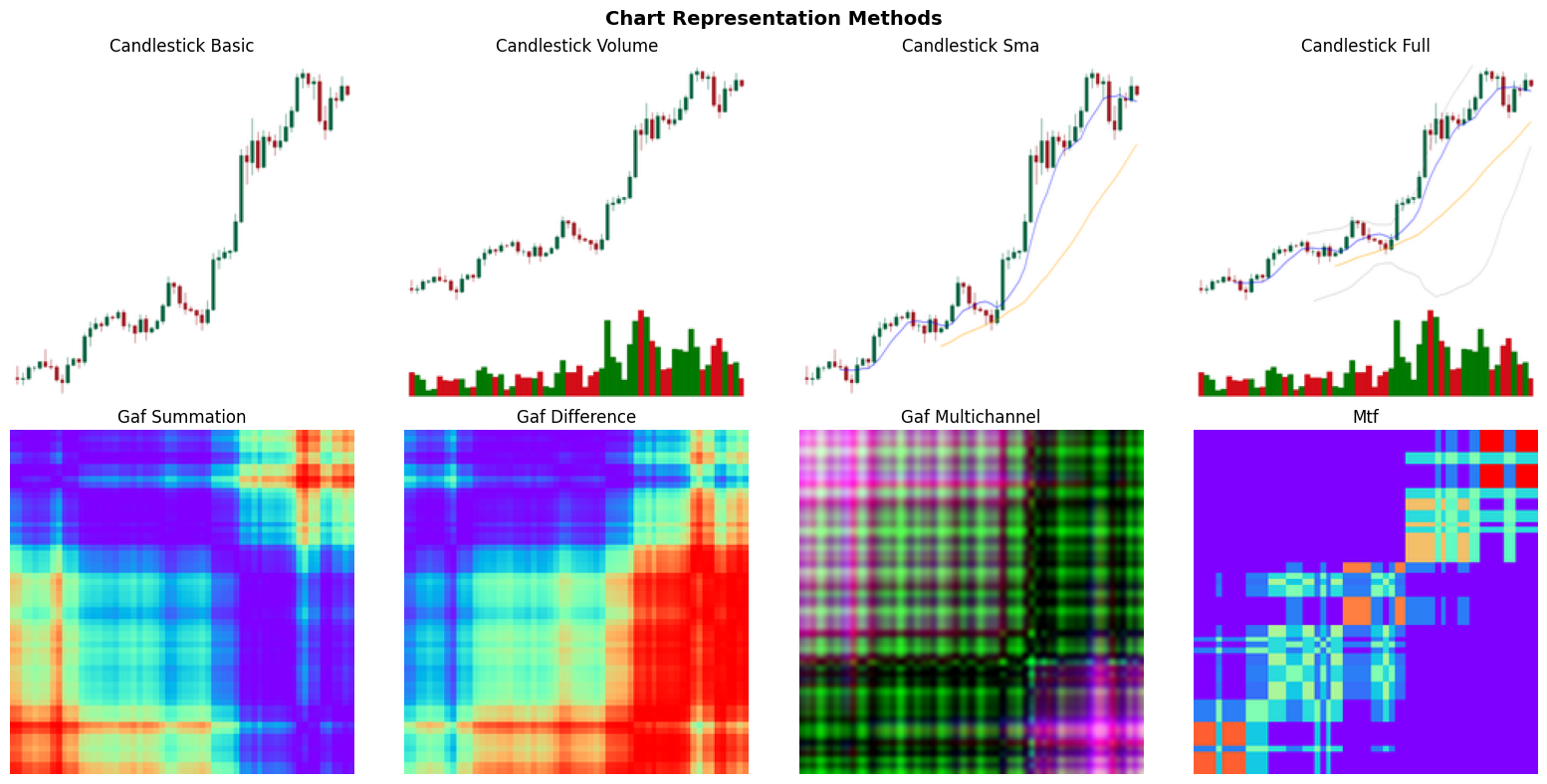}
\caption{Chart representation methods. Top row: candlestick variants (basic, volume, SMA, full). Bottom row: mathematical encodings (GAF summation, difference, multi-channel, MTF).}
\label{fig:chart_methods}
\end{figure}

Fig.~\ref{fig:price_history} displays the price history for all three assets, illustrating the high volatility characteristic of cryptocurrency markets compared to traditional equities. Fig.~\ref{fig:chart_methods} shows examples of the different chart representation methods evaluated in this study.

\subsection{Experiment 2: Effect of Image Encoding}

Table~\ref{tab:encoding} compares the three image encoding methods. Raw candlestick charts achieve an AUC-ROC of 0.760, substantially outperforming both GAF variants. Surprisingly, the GAF methods achieved AUC scores below 0.5 (0.310 and 0.252), indicating that their predictions are inversely correlated with the true labels. This counterintuitive result suggests that the mathematical encoding loses important visual information that CNNs can extract from traditional chart representations. The high accuracy and F1 scores for GAF Multi-channel are misleading artifacts of threshold selection; the AUC-ROC reveals the true discrimination ability.

\begin{table}[h]
\centering
\caption{Comparison of Image Encoding Methods}
\label{tab:encoding}
\begin{tabular}{lccc}
\toprule
\textbf{Encoding} & \textbf{Acc} & \textbf{F1} & \textbf{AUC} \\
\midrule
\textbf{Raw Candlestick} & 49.3\% & 0.367 & \textbf{0.760} \\
GAF (Summation) & 33.3\% & 0.419 & 0.310 \\
GAF (Multi-channel) & 64.0\% & 0.780 & 0.252 \\
\bottomrule
\end{tabular}
\end{table}

\begin{figure}[h]
\centering
\includegraphics[width=\columnwidth]{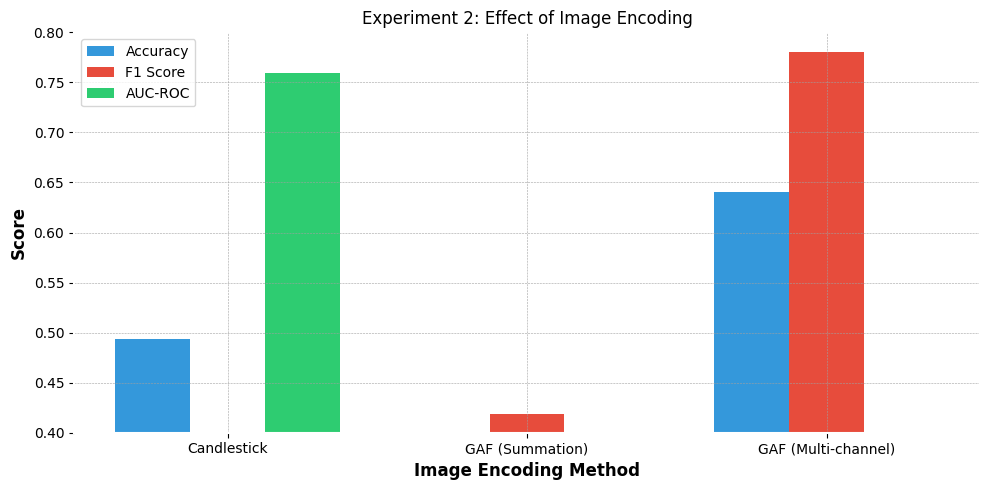}
\caption{Experiment 2: Image encoding comparison. Raw candlestick charts achieve significantly higher AUC-ROC than GAF encodings.}
\label{fig:exp2_encoding}
\end{figure}

\subsection{Experiment 3: Effect of Chart Components}

Table~\ref{tab:components} reveals a surprising finding: simpler charts perform better. Price-only candlesticks achieve the highest AUC-ROC of 0.815, while adding volume bars reduces performance to 0.706. Adding moving averages further degrades AUC to 0.362, and including all indicators yields 0.704. These results suggest that additional visual elements distract the CNN from learning core price patterns, or that the information in technical indicators is redundant given the price data already present in the candlesticks.

\begin{table}[h]
\centering
\caption{Impact of Visual Chart Components}
\label{tab:components}
\begin{tabular}{lccc}
\toprule
\textbf{Components} & \textbf{Acc} & \textbf{F1} & \textbf{AUC} \\
\midrule
\textbf{Price Only} & 53.3\% & 0.444 & \textbf{0.815} \\
+ Volume & 52.0\% & 0.419 & 0.706 \\
+ SMA & 50.7\% & 0.673 & 0.362 \\
+ Volume + SMA & 50.7\% & 0.393 & 0.703 \\
+ All (Vol+SMA+BB) & 48.0\% & 0.339 & 0.704 \\
\bottomrule
\end{tabular}
\end{table}

\begin{figure}[h]
\centering
\includegraphics[width=\columnwidth]{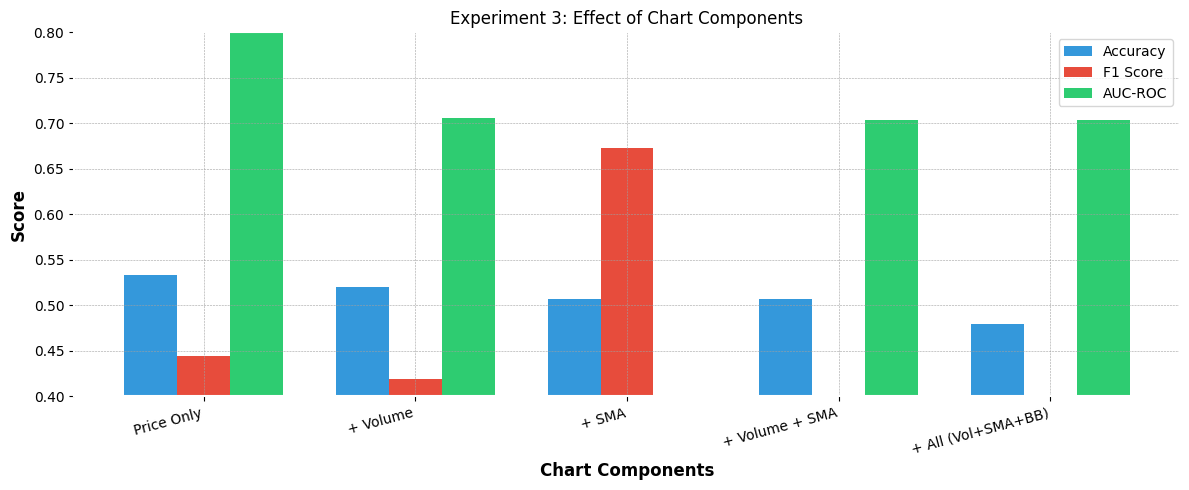}
\caption{Experiment 3: Chart components impact. Price-only candlesticks achieve highest AUC-ROC, with performance degrading as indicators are added.}
\label{fig:exp3_components}
\end{figure}

\subsection{Experiment 4: Effect of Lookback Window}

Table~\ref{tab:lookback} shows that a 30-day lookback window achieves the highest AUC-ROC of 0.734. Shorter windows of 14 days yield only 0.551 AUC, likely because they lack sufficient historical context to identify meaningful patterns. Conversely, longer windows of 60 and 90 days produce AUC scores of 0.401 and 0.559 respectively, suggesting that older price patterns introduce noise that obscures more recent, relevant signals. The 30-day window aligns with common trading practice, where monthly patterns are often considered most relevant for short-term prediction.

\begin{table}[h]
\centering
\caption{Impact of Historical Lookback Period}
\label{tab:lookback}
\begin{tabular}{lccc}
\toprule
\textbf{Lookback (days)} & \textbf{Acc} & \textbf{F1} & \textbf{AUC} \\
\midrule
14 & 57.3\% & 0.680 & 0.551 \\
\textbf{30} & 53.3\% & 0.444 & \textbf{0.734} \\
60 & 60.0\% & 0.750 & 0.401 \\
90 & 50.7\% & 0.667 & 0.559 \\
\bottomrule
\end{tabular}
\end{table}

\subsection{Experiment 5: Effect of Image Resolution}

Table~\ref{tab:resolution} demonstrates that 128$\times$128 resolution achieves the highest AUC-ROC of 0.892, representing our best overall result. Lower resolution of 64$\times$64 yields 0.713 AUC, likely because fine-grained candlestick patterns are lost through downsampling. Higher resolution of 224$\times$224 achieves 0.849 AUC, slightly below the optimal, possibly because the additional pixels introduce noise or require more training data to learn effectively. The 128$\times$128 resolution appears to represent a sweet spot that preserves sufficient visual detail while avoiding unnecessary complexity.

\begin{table}[h]
\centering
\caption{Impact of Image Resolution}
\label{tab:resolution}
\begin{tabular}{lccc}
\toprule
\textbf{Resolution} & \textbf{Acc} & \textbf{F1} & \textbf{AUC} \\
\midrule
64$\times$64 & 58.7\% & 0.617 & 0.713 \\
\textbf{128$\times$128} & 46.7\% & 0.310 & \textbf{0.892} \\
224$\times$224 & 45.3\% & 0.281 & 0.849 \\
\bottomrule
\end{tabular}
\end{table}

\subsection{Experiment 6: Effect of Architecture}

Table~\ref{tab:architecture} presents perhaps our most surprising finding: the simple 4-layer CNN with only 422,000 parameters outperforms all larger pretrained models on AUC-ROC. The simple CNN achieves 0.831 AUC, compared to 0.662 for ResNet18 (11.2M parameters), 0.365 for EfficientNet-B0 (4.0M parameters), and 0.570 for ViT-Tiny (5.5M parameters). The larger models show signs of severe overfitting, achieving near-zero training loss but poor test discrimination. This suggests that larger models require substantially more training data than our 500 samples provide, or that features learned from ImageNet do not transfer as effectively to financial charts as to other visual domains.

\begin{table}[h]
\centering
\caption{Comparison of Model Architectures}
\label{tab:architecture}
\begin{tabular}{lcccc}
\toprule
\textbf{Model} & \textbf{Params} & \textbf{Acc} & \textbf{F1} & \textbf{AUC} \\
\midrule
\textbf{Simple CNN} & 422K & 44.0\% & 0.250 & \textbf{0.831} \\
ResNet18 & 11.2M & 60.0\% & 0.681 & 0.662 \\
EfficientNet & 4.0M & 64.0\% & 0.777 & 0.365 \\
ViT-Tiny & 5.5M & 66.7\% & 0.797 & 0.570 \\
\bottomrule
\end{tabular}
\end{table}

\subsection{Experiment 7: Effect of Transfer Learning}

Despite the challenges faced by larger models, Table~\ref{tab:transfer} shows that ImageNet pretraining consistently improves performance compared to random initialization. For ResNet18, pretraining improves AUC-ROC from 0.654 to 0.695, a gain of 4.1 percentage points. For EfficientNet-B0, the improvement is even more substantial, from 0.372 to 0.533, representing a 16.1 percentage point increase. These results suggest that despite the significant visual domain gap between natural images and financial charts, low-level features learned from ImageNet---such as edge detectors and texture recognizers---transfer meaningfully to chart analysis.

\begin{table}[h]
\centering
\caption{Transfer Learning Impact}
\label{tab:transfer}
\begin{tabular}{llccc}
\toprule
\textbf{Model} & \textbf{Init} & \textbf{Acc} & \textbf{F1} & \textbf{AUC} \\
\midrule
\multirow{2}{*}{ResNet18} & Scratch & 41.3\% & 0.185 & 0.654 \\
 & \textbf{ImageNet} & 57.3\% & 0.719 & \textbf{0.695} \\
\midrule
\multirow{2}{*}{EfficientNet} & Scratch & 65.3\% & 0.790 & 0.372 \\
 & \textbf{ImageNet} & 54.7\% & 0.660 & \textbf{0.533} \\
\bottomrule
\end{tabular}
\end{table}

\subsection{Experiment 8: Generalization Across Assets}

Table~\ref{tab:assets} reveals substantial variation in model performance across different assets. Bitcoin shows the strongest predictability with an AUC-ROC of 0.734, while Ethereum achieves only 0.492 (near random chance) and the S\&P 500 yields 0.459 (below random chance). These results suggest that visual patterns in Bitcoin charts may be more pronounced or consistent than in other assets, possibly due to the dominance of retail traders who rely heavily on technical analysis. The poor performance on traditional equities aligns with efficient market hypothesis predictions that liquid markets should exhibit less predictability.

\begin{table}[h]
\centering
\caption{Cross-Asset Performance}
\label{tab:assets}
\begin{tabular}{lcccc}
\toprule
\textbf{Asset} & \textbf{Samples} & \textbf{Acc} & \textbf{F1} & \textbf{AUC} \\
\midrule
\textbf{Bitcoin} & 500 & 69.3\% & 0.793 & \textbf{0.734} \\
Ethereum & 500 & 61.3\% & 0.760 & 0.492 \\
S\&P 500 & 500 & 33.3\% & 0.286 & 0.459 \\
\bottomrule
\end{tabular}
\end{table}

\section{Interpretability Analysis}

\subsection{GradCAM Visualization}

To understand what visual features the CNN learns, we apply Gradient-weighted Class Activation Mapping (GradCAM) \cite{selvaraju2017grad} to visualize regions of the chart that most influence predictions. GradCAM computes the gradient of the output with respect to the final convolutional layer activations, then uses these gradients to weight the activation maps. The resulting heatmap highlights regions that contribute most to the classification decision, providing interpretable insights into model behavior.

\begin{figure}[h]
\centering
\includegraphics[width=\columnwidth]{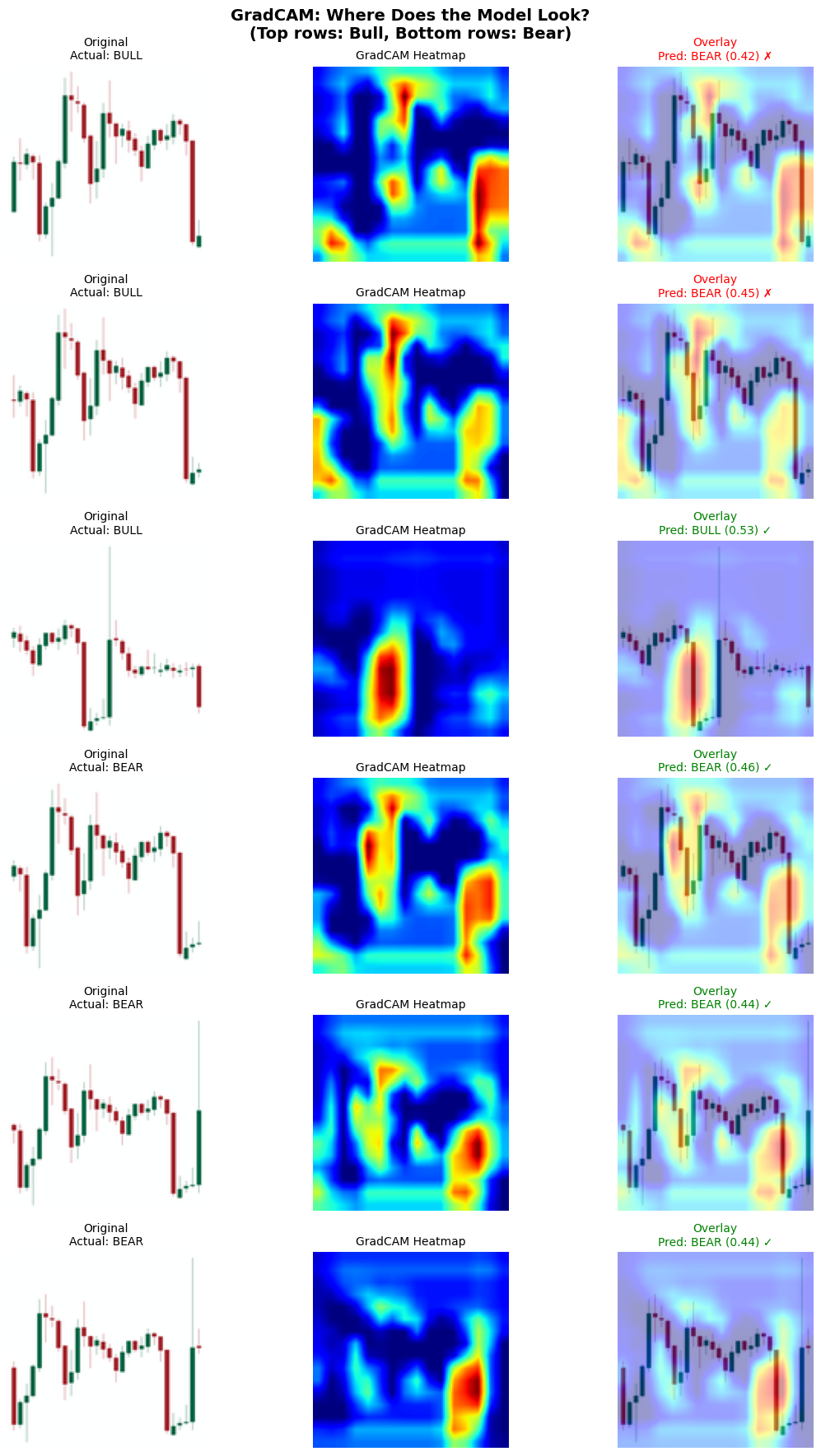}
\caption{GradCAM visualization showing model attention for bull (top rows) and bear (bottom rows) samples. Left: original chart. Center: attention heatmap. Right: overlay with prediction. The model focuses on recent price action and large candles.}
\label{fig:gradcam}
\end{figure}

Fig.~\ref{fig:gradcam} displays GradCAM visualizations for several test samples, showing the original chart, the attention heatmap, and an overlay combining both. The visualizations reveal consistent patterns in model attention that align with intuitive trading concepts.

\subsection{Key Observations}

Our GradCAM analysis reveals several interpretable patterns in how the model processes chart images. Most prominently, the model consistently attends to the right side of charts, corresponding to the most recent candles. This suggests the network has learned that recent momentum is most predictive of near-term regime, a principle well-established in technical analysis.

Additionally, high activation regions frequently correspond to candles with large bodies, indicating significant price movements within a single day. The model appears to recognize the importance of momentum signals encoded in these prominent visual features. At potential reversal points where sequences of green candles transition to red or vice versa, the model shows increased attention, aligning with classical technical analysis concepts of trend reversal detection.

We also observe that confident predictions tend to produce focused activation maps concentrated on specific chart regions, while uncertain predictions exhibit more diffuse attention patterns spread across the entire image. This behavior mirrors how human traders might focus intently on clear signals while being uncertain when no obvious patterns emerge.

\subsection{Alignment with Technical Analysis}

These observations suggest the CNN learns features consistent with how human traders analyze charts. The attention to recent large candles parallels momentum-based trading strategies. The focus on candle color sequences reflects trend identification approaches. The increased attention at color transitions mirrors reversal pattern recognition used by technical analysts. This alignment provides confidence that the model learns meaningful visual features rather than spurious correlations in the training data.

\begin{figure}[h]
\centering
\includegraphics[width=\columnwidth]{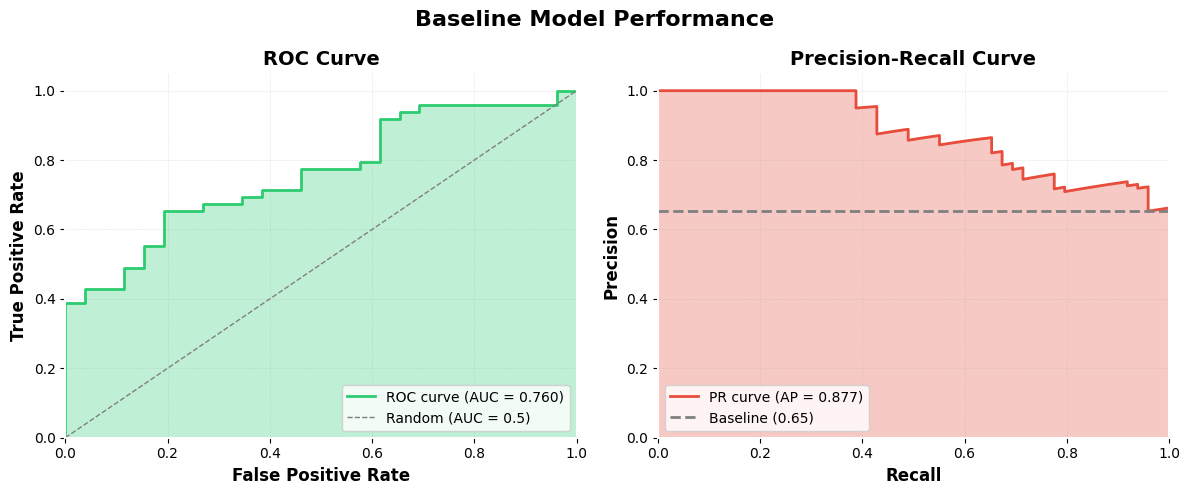}
\caption{ROC curve (left) and Precision-Recall curve (right) for baseline model. AUC-ROC of 0.760 demonstrates discrimination above random chance.}
\label{fig:roc_pr}
\end{figure}

\begin{figure}[h]
\centering
\includegraphics[width=0.75\columnwidth]{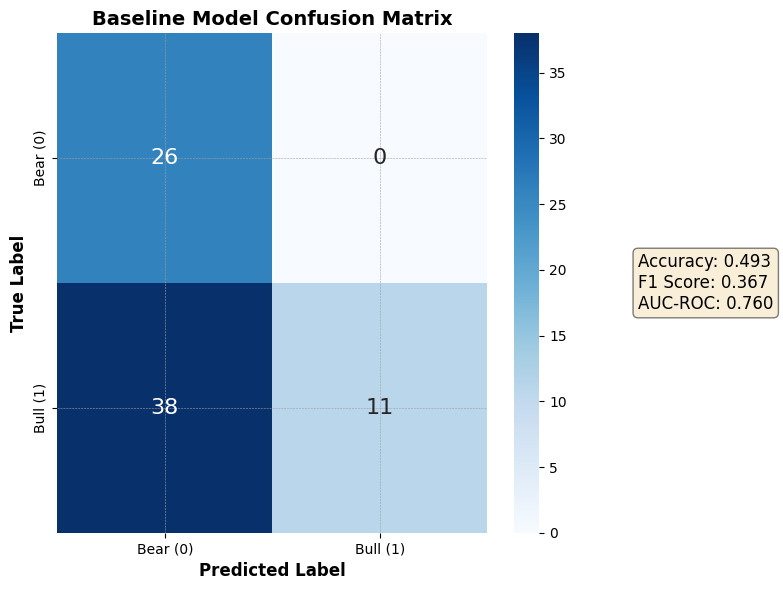}
\caption{Confusion matrix for baseline model. The model correctly identifies bear regimes but struggles with bull regime recall.}
\label{fig:confusion}
\end{figure}

\section{Discussion}

\subsection{Summary of Key Findings}

Our systematic experimental study yields five principal findings that inform the design of visual trading systems.

First, regarding image encoding, raw candlestick charts significantly outperform mathematically-derived encodings such as Gramian Angular Fields. Traditional visualizations preserve predictive information better than abstract transformations, likely because they maintain the spatial relationships that CNNs are designed to exploit. The poor performance of GAF encodings, with AUC scores below random chance, suggests that the mathematical transformation destroys rather than preserves relevant patterns.

Second, concerning chart components, simpler representations consistently outperform more complex alternatives. Adding technical indicators such as volume bars, moving averages, and Bollinger Bands degrades rather than improves performance. This counterintuitive result suggests that visual clutter distracts the CNN from learning core price patterns, or that the information in these indicators is already implicitly encoded in the candlestick visualization.

Third, with respect to model architecture, a simple 4-layer CNN with 422,000 parameters outperforms models with 10 to 25 times more parameters. The larger pretrained models exhibit severe overfitting with our limited training data of 500 samples. This finding has important practical implications: practitioners should start with simple models rather than assuming that larger, more sophisticated architectures will perform better.

Fourth, transfer learning from ImageNet provides consistent improvements of 4 to 16 percentage points in AUC-ROC, despite the substantial visual domain gap between natural images and financial charts. Low-level features such as edge detection and texture recognition appear to transfer meaningfully across these disparate domains.

Fifth, regarding generalization, Bitcoin exhibits substantially stronger visual predictability than Ethereum or traditional equities. This may reflect the dominance of retail traders in Bitcoin markets who rely heavily on technical analysis, creating self-fulfilling patterns that CNNs can learn to recognize.

\subsection{Practical Implications}

For practitioners seeking to apply deep learning to visual chart analysis, our results suggest several guidelines. Raw candlestick charts should be preferred over mathematical encodings, and charts should be kept simple without cluttering them with technical indicators. Development should begin with small models, as larger architectures are not better with limited training data. Based on our experiments, 30-day lookback windows and 128$\times$128 resolution represent good starting points. Transfer learning should be considered even for this specialized visual domain, as ImageNet pretraining provides meaningful benefits. Finally, better results should be expected on cryptocurrency markets than on traditional equities, potentially due to differences in market microstructure and participant behavior.

\subsection{Limitations}

Several limitations qualify our findings and suggest caution in their application. With 500 samples per experiment, results may exhibit high variance, and production systems should use substantially larger datasets. Predictable patterns may be arbitraged away over time as markets become more efficient, potentially degrading model performance. Our evaluation focuses on discrimination ability rather than risk-adjusted returns after accounting for transaction costs and slippage. The patterns identified may be regime-dependent and could degrade as market conditions change. Finally, our analysis uses only daily data; intraday or weekly patterns may differ substantially and warrant separate investigation.

\subsection{Future Work}

Several extensions could build upon this work. Multi-timeframe fusion combining daily and hourly charts might capture patterns at different scales. Attention-based architectures designed specifically for financial charts could improve upon generic vision models. Integration with fundamental and sentiment data might provide complementary signals. Full backtesting with realistic transaction costs would assess practical trading viability. Most importantly, substantially larger datasets would enable effective use of bigger models and provide more reliable performance estimates.

\begin{figure*}[b]
\centering
\includegraphics[width=0.95\textwidth]{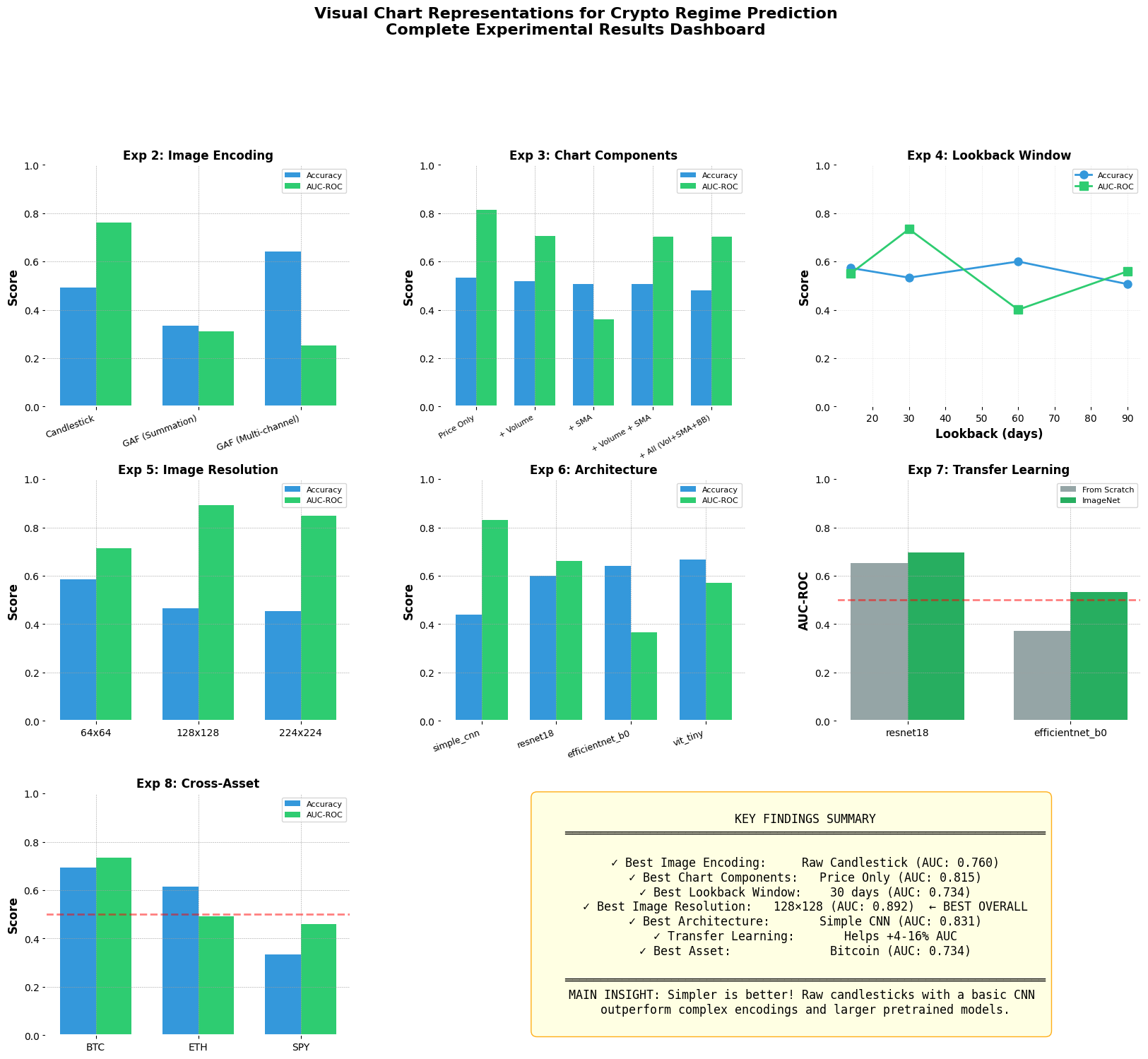}
\caption{Complete experimental results dashboard summarizing all eight experiments. Each subplot shows accuracy (blue) and AUC-ROC (green) for different configurations. Key finding: simpler representations consistently outperform complex alternatives.}
\label{fig:dashboard}
\end{figure*}

\section{Conclusion}

This paper presented a systematic study of visual chart representations for cryptocurrency regime prediction through eight controlled experiments evaluating image encoding methods, chart components, neural network architectures, and transfer learning strategies.

Our investigation yields several actionable conclusions. Simplicity consistently wins: raw candlestick charts with a basic CNN achieve the best discrimination with an AUC-ROC of 0.892, outperforming complex encodings and large pretrained models. Less is more: price-only charts outperform indicator-laden alternatives, and 128$\times$128 resolution beats 224$\times$224. Transfer learning helps: despite the domain gap, ImageNet pretraining improves performance by 4 to 16 percentage points. Asset selection matters: Bitcoin shows stronger visual predictability than Ethereum or the S\&P 500. Finally, interpretable features emerge: GradCAM analysis confirms the model learns meaningful visual patterns aligned with technical analysis principles, focusing on recent price action and significant momentum signals.

These results provide practical guidance for practitioners seeking to apply deep learning to visual chart analysis. The consistent finding that simpler approaches outperform complex alternatives has important implications for the design of visual trading systems, suggesting that engineering effort should focus on data quality and problem formulation rather than model complexity.

\subsection*{Code Availability}

All code, trained models, and experimental results are available at: \url{https://github.com/dustinhaggett/chart-vision-regime}

\end{document}